\documentclass[10pt,twocolumn,letterpaper]{article}

\usepackage[pagenumbers]{cvpr} 

\usepackage{graphicx}
\usepackage{amsmath}
\usepackage{amssymb}
\usepackage{booktabs}
\usepackage{graphicx}
\usepackage{multirow}
\usepackage[table]{xcolor}
\usepackage{subcaption}
\usepackage{bbding}

\usepackage[pagebackref,breaklinks,colorlinks]{hyperref}

\usepackage[capitalize]{cleveref}
\crefname{section}{Sec.}{Secs.}
\Crefname{section}{Section}{Sections}
\Crefname{table}{Table}{Tables}
\crefname{table}{Tab.}{Tabs.}

\begin{document}

\title{Lawin Transformer: Improving Semantic Segmentation Transformer with Multi-Scale Representations via Large Window Attention}

\author{Haotian Yan, Chuang Zhang and Ming Wu \\
Pattern Recognition and Intelligent System Lab, \\ Beijing University of Posts and Telecommunications\\
{\tt\small \{yanhaotian, zhangchuang, wuming\}@bupt.edu.cn}
}
\maketitle

\begin{abstract}
Multi-scale representations are crucial for semantic segmentation. The community has witnessed the flourish of semantic segmentation convolutional neural networks (CNN) exploiting multi-scale contextual information. Motivated by that the vision transformer (ViT) is powerful in image classification, some semantic segmentation ViTs are recently proposed, most of them attaining impressive results but at a cost of computational economy. In this paper, we succeed in introducing multi-scale representations into semantic segmentation ViT via window attention mechanism and further improves the performance and efficiency. To this end, we introduce large window attention which allows the local window to query a larger area of context window at only a little computation overhead. By regulating the ratio of the context area to the query area, we enable the \textit{large window attention} to capture the contextual information at multiple scales. Moreover, the framework of spatial pyramid pooling is adopted to collaborate with \textit{the large window attention}, which presents a novel decoder named \textbf{la}rge \textbf{win}dow attention spatial pyramid pooling (LawinASPP) for semantic segmentation ViT. Our resulting ViT, Lawin Transformer, is composed of an efficient hierachical vision transformer (HVT) as encoder and a LawinASPP as decoder. The empirical results demonstrate that Lawin Transformer offers an improved efficiency compared to the existing method. Lawin Transformer further sets new state-of-the-art performance on Cityscapes (84.4\% mIoU), ADE20K (56.2\% mIoU) and COCO-Stuff datasets. The code will be released at \href{https://github.com/yan-hao-tian/lawin}{https://github.com/yan-hao-tian/lawin}.

\end{abstract}

\section{Introduction}
\label{sec:intro}

\begin{figure}[t]
\begin{center}
  \includegraphics[width=0.5\textwidth]{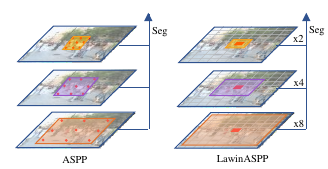} \vspace{-0.8cm}
\end{center}
  \caption{Difference between LawinASPP and ASPP. In ASPP, atrous convolution with different dilation rates captures representations at multple scalees. In contrast, LawinASPP replaces atrous convolution with our proposed \textit{large window attention}. The red window represents the query area. The yellow, orange and purple windows represent the context area with different spatial sizes.} 
\label{fig:LawinASPP}
\end{figure}

Semantic segmentation is one of the most significant dense prediction tasks in computer vision. With the prosperity of deep convolutional neural network (CNN) in this field, the CNN-based semantic segmentation pipeline gains more and more popularity in a wide range of practical applications such as self-driving cars, medical imaging analysis and remote sensing imagery interpretation~\cite{ronneberger2015u, mnih2010learning, siam2017deep}. Having scrutinized the famed semantic segmentation CNN, we note that a series of work largely focus on exploiting multi-scale representations~\cite{he2019adaptive, he2019dynamic, chen2017rethinking, zhao2017pyramid, chen2018encoder, yuan2018ocnet}, which plays a vital role of understanding the context prior at multiple scales. To incorporate the rich contextual information, most of these works apply filters or pooling operations, such as atrous convolution~\cite{yu2015multi} and adaptive pooling, to the spatial pyramid pooling (SPP) module~\cite{grauman2005pyramid, lazebnik2006beyond}.

Since the impressive performance of Vision Transformer (ViT) on image classification~\cite{touvron2021training, DBLP:conf/iclr/DosovitskiyB0WZ21}, there are some efforts to resolve semantic segmentation with pure transformer models, still outperforming the previous semantic segmentation CNN by a large margin~\cite{liu2021swin, strudel2021segmenter, zheng2021rethinking, xiao2018unified}. However, it takes a very high computation cost to implement these semantic segmentation ViTs, especially when the input image is large. In order to tackle this issue, the method purely based on the hierarchical vision transformer (HVT) has emerged with saving much computational budget. Swin Transformer is one of the most representative HVTs achieving state-of-the-art results on many vision tasks~\cite{liu2021swin}, whilst it employs a heavy decoder~\cite{xiao2018unified} to classify pixels. SegFormer refines the design of both encoder and decoder, resulting a very efficient semantic segmentation ViT~\cite{xie2021segformer}. But it is problematic that SegFormer solely relies on increasing the model capacity of encoder to progressively improve performance, which has a potentially lower efficiency ceiling. 

Through the above analysis, we think one major problem for current semantic segmentation ViT is lack of multi-scale contextual information, thus impairing the performance and efficiency. To overcome the limitation, we present a novel window attention mechanism named \textit{large window attention}. In large window attention, the uniformly split patch queries the context patch covering a much larger region as illustrated in Fig~\ref{fig:LawinASPP}, whereas the patch in \textit{local window attention} merely queries itself. On the other hand, considering that the attention would become computationally prohibitive with the enlargement of context patch, we devise a simple yet effective strategy to alleviate the \textit{dilemma of large context}. Specifically, we first pool the large context patch to spatial dimension of the corresponding query patch in order to preserve the original computational complexity. Then we enable the multi-head mechanism in large window attention and set the number of head strictly equal to the square of the downsampling ratio $R$ while pooling the context, mainly for recovering the discarded dependencies between query and context. Finally, inspired by token-mixing MLP in MLP-Mixer~\cite{tolstikhin2021mlp}, we apply $R^2$ \textit{position-mixing} operations on the $R^2$ subspaces of head respectively, strengthening the spatially representational power of multi-head attention. Therefore, the patch in our proposed large window attention can capture contextual information at any scales, merely yielding a little computational overhead caused by \textit{position-mixing} operations. Coupled with large window attention with different ratios $R$, a SPP module evolves into a \textit{large window attention spatial pyramid pooling} (LawinASPP), which one can employ like ASPP (Atrous Spatial Pyramid Pooling)~\cite{chen2017rethinking} and PPM (Pyramid Pooling Module)~\cite{zhao2017pyramid} to exploit multi-scale representations for semantic segmentation.

We extend the efficient HVT to Lawin Transformer by placing the LawinASPP at the top of HVT, which introduces the multi-scale representations into the semantic segmentation ViT. The performance and efficiency of Lawin Transformer is evaluated on Cityscapes~\cite{cordts2016cityscapes}, ADE20K~\cite{zhou2017scene} and COCO-Stuff~\cite{caesar2018coco} datasets. We conduct extensive experiments to compare Lawin Transformer with existing HVT-based semantic segmentation method~\cite{liu2021swin, xie2021segformer, cheng2021maskformer}. An improved efficiency of Lawin Transformer is proved by that Lawin Transformer spends less computational resource in attaining the better performance. Besides, our experiments show that Lawin Transformer outperforms other state-of-the-art methods consistently on these benchmarks. 
\begin{figure*}[t]
\begin{center}
   \includegraphics[width=1.0\textwidth]{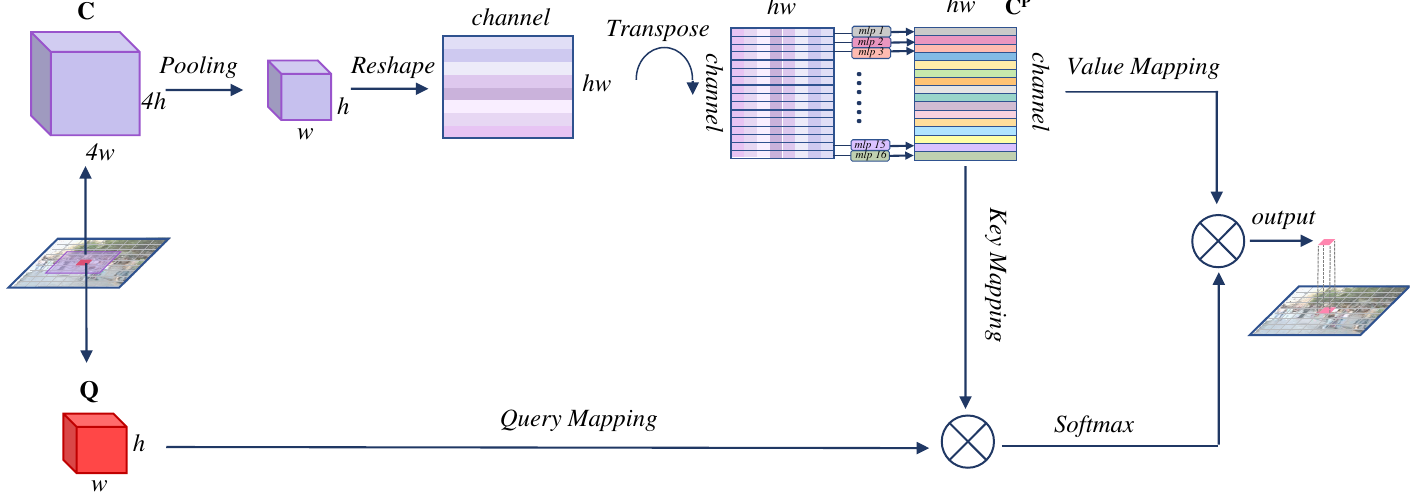}
\end{center}
   \caption{A large window attention. The red patch $\textbf{Q}$ is the query patch and the purple patch $\textbf{C}$ is the context patch. The context is reshaped and fed into token-mixing MLPs. The output context $\textbf{C}^\textbf{P}$ is named \textit{position-mixed context}. Best viewed in color. } \vspace{-0.2cm}
\label{fig:lawin attention}
\end{figure*}

\section{Related Work}
\label{sec:2}

\subsection{Semantic Segmentation}
Semantic segmentation models based on fully convolutional neural network (FCN)~\cite{long2015fully} are the most promising ways to accomplish the pixel-level classification. Towards precise scene understanding, consecutive improvements have been developed for semantic segmentation CNN in many aspects. \cite{amirul2017gated, lin2017refinenet, badrinarayanan2017segnet, ronneberger2015u} mitigates the boundary information shortage of high-level feature. \cite{chen2017deeplab, peng2017large, chen2014semantic, yu2015multi} are proposed to enlarge the receptive field of model. The spatial pyramid pooling (SPP) module has proved to be effective in exploiting multi-scale representations, which gathers scene clues from local context to global context~\cite{chen2018encoder, zhao2017pyramid, he2019adaptive, he2019dynamic}. An alternative line of work utilizes the variants of self-attention mechanism to model dependencies among representations~\cite{cao2019gcnet, huang2019ccnet, wang2018non, li2018pyramid, zhao2018psanet, yu2020context, fu2019dual, yin2020disentangled}. 

\subsection{Vision Transformer}
Transformer has revolutionized neural language processing and proven extremely successful in computer vision. ViT~\cite{DBLP:conf/iclr/DosovitskiyB0WZ21} is the first end-to-end Vision Transformer for image classification by projecting the input image into a token sequence and attach it to a class token. DeiT~\cite{touvron2021training} improves the data efficiency of training ViT with a token distillation pipeline. Apart from the sequence-to-sequence structure, the efficiency of PVT~\cite{wang2021pyramid} and Swin Transformer~\cite{liu2021swin} sparks much interests in exploring the Hierarchical Vision Transformer (HVT)~\cite{graham2021levit, wu2021cvt, chu2021twins, yan2021contnet}. ViT is also extended to solve the low-level tasks and dense prediction problems~\cite{carion2020end, arnab2021vivit, esser2021taming}. Specially, concurrent semantic segmentation methods driven by ViT presents impressive performance. SETR~\cite{zheng2021rethinking} deploys the ViT as an encoder and upsamples the output patch embedding to classify pixels. Swin Transformer extends itself to a semantic segmentation ViT by connecting a UperNet~\cite{xiao2018unified}. Segmenter~\cite{strudel2021segmenter} depends on the ViT/DeiT as backbone and propose a mask Transformer decoder. Segformer~\cite{xie2021segformer} shows a simple, efficient yet powerful design of encoder and decoder for semantic segmentation. MaskFormer~\cite{cheng2021maskformer} reformulates the semantic segmentation as a mask classification problem, having much fewer FLOPs and parameters compared to Swin-UperNet. In this paper, we take a new step towards a more efficient design of semantic segmentation ViT, by introducing multi-scale representations into the HVT. 

\subsection{MLP-Mixer}
MLP-Mixer~\cite{tolstikhin2021mlp} is a novel neural network much simpler than ViT. Similar to ViT, MLP-Mixer first adopts a linear projection to obtain a token sequence like ViT. The sharp dinstinction is that MLP-Mixer is entirely based on multi-layer perceptrons (MLP), because it replaces the self-attention in transformer layer with the token-mixing MLP. Token-Mixing MLP acts along the channel dimension, mixing the token (position) to learn spatial representations. In our proposed large window attention, token-mixing MLP is applied to the pooled context patch, which we call \textit{position-mixing} to boost the spatial representations of multi-head attention.




\section{Method}
In this part, we first briefly introduce multi-head attention and token-mixing MLP. Then we elaborate large window attention and describe the architecture of LawinASPP. Finally, the overall structure of Lawin Transformer is presented.

\subsection{Background}
\label{sec:3.1}
Multi-head attention is the core of Transformer layer. In the Hierarchical Vision Transformer (HVT), the operation of multi-head attention is limited to local uniformly split window, which is called local window attention. Assuming the input is a 2D feature map denoted as $ \mathbf{x_{\rm{2d}}} \in \mathbb{R} ^ \mathit{C \times H \times W}$, we can formulate the action of window attention as:  

\begin{align}
& \mathbf{\hat{x}_{\rm{2d}}} = \rm{Reshape}\left(\mathit{h},\mathit{\frac{HW}{P^2}} , \mathit{\frac{C}{h}}, \mathit{P}, \mathit{P} \right)\left( \mathbf{x_{\rm{2d}}}\right),  \label{eq:rel1} \\
& \mathbf{\mathbf{x_{\rm{2d}}}} = \rm{Reshape}\left( \mathit{C}, \mathit{H}, \mathit{W}   \right)\left( \rm{MHA}\left(\mathbf{\hat{x}_{\rm{2d}}}\right)\right)+x_{\rm{2d}},  \label{eq:rel2}
\end{align}
where $h$ is the head number and $P$ is the spatial size of windows, and $\rm{MHA} \left( \space \right)$ is the Multi-Head Attention (MHA) mechanism. The basic operation of MHA can be described as:
\begin{align}
& \mathit{A}=\rm{softmax} \left( \frac{ {(\mathbf{W_{\rm{q}}}\mathbf{x_{\rm{2d}}})}{(\mathbf{W_{\rm{k}}}{\mathbf{x_{\rm{2d}}})}^{\rm{T}}}}{ \sqrt{\mathit{D_{h}}}}  \right) \left(\mathbf{W_{\rm{v}}}\mathbf{x_{\rm{2d}}}\right),  
\label{eq:rel3} \\
& \rm{MHA}=\rm{concat}\left[\mathit{A_{1}};\mathit{A_{2}};...;\mathit{A_{h}}\right] {\mathbf{W_{\rm{mha}}}}, \label{eq:rel4}
\end{align}
where $\mathbf{W_{\rm{q}}}$, $\mathbf{W_{\rm{k}}} $ and $\mathbf{W_{\rm{v}}}  \in \mathbb{R} ^ \mathit{C\times D_h}$ are the learned linear transformations and $\mathbf{W_{\rm{mhsa}}}  \in \mathbb{R} ^ \mathit{D\times C}$ is the learned weights that aggregates multiple attention values. $D_h$ is typically set to $D \slash h$ and $D$ is the embedding dimension.

Token-mixing MLP is the core of MLP-Mixer which can aggregate spatial information, by allowing the spatial position to communicate each other. Given the input 2D feature map $ \mathbf{x_{\rm{2d}}} \in \mathbb{R} ^ \mathit{C \times  H \times W  }$, the operation of token-mixing MLP can be formulated as:
\begin{align}
& \mathbf{\hat{x}_{\rm{2d}}} = \rm{Reshape}\left(\mathit{C}, \mathit{HW} \right)\left( \mathbf{x_{\rm{2d}}}\right),  \label{eq:rel5} \\
& \mathbf{x_{\rm{2d}}} = \rm{Reshape}\left(\mathit{C}, \mathit{H}, \mathit{W} \right)\left( MLP\left(\mathbf{\hat{x}_{\rm{2d}}}\right)\right)+\mathbf{x_{\rm{2d}}}, \label{eq:rel6}\\
&  \rm{MLP} \left(\mathbf{x_{\rm{2d}}} \right) =\mathbf{W_{\rm{2}}} \sigma \left( \mathbf{W_{\rm{1}}} \mathbf{x_{\rm{2d}}} \right),  \label{eq:rel7}
\end{align}
where $\mathbf{W_{\rm{1}}} \in \mathbb{R} ^ \mathit{HW\times D_{mlp}}$ and $\mathbf{W_{\rm{2}}} \in \mathbb{R} ^ \mathit{ D_{mlp} \times HW}$ are both learned linear transformations, and $\sigma$ is the activation function providing non-linearity.

\begin{figure*}[t]
\begin{center}
   \includegraphics[width=1.0\textwidth]{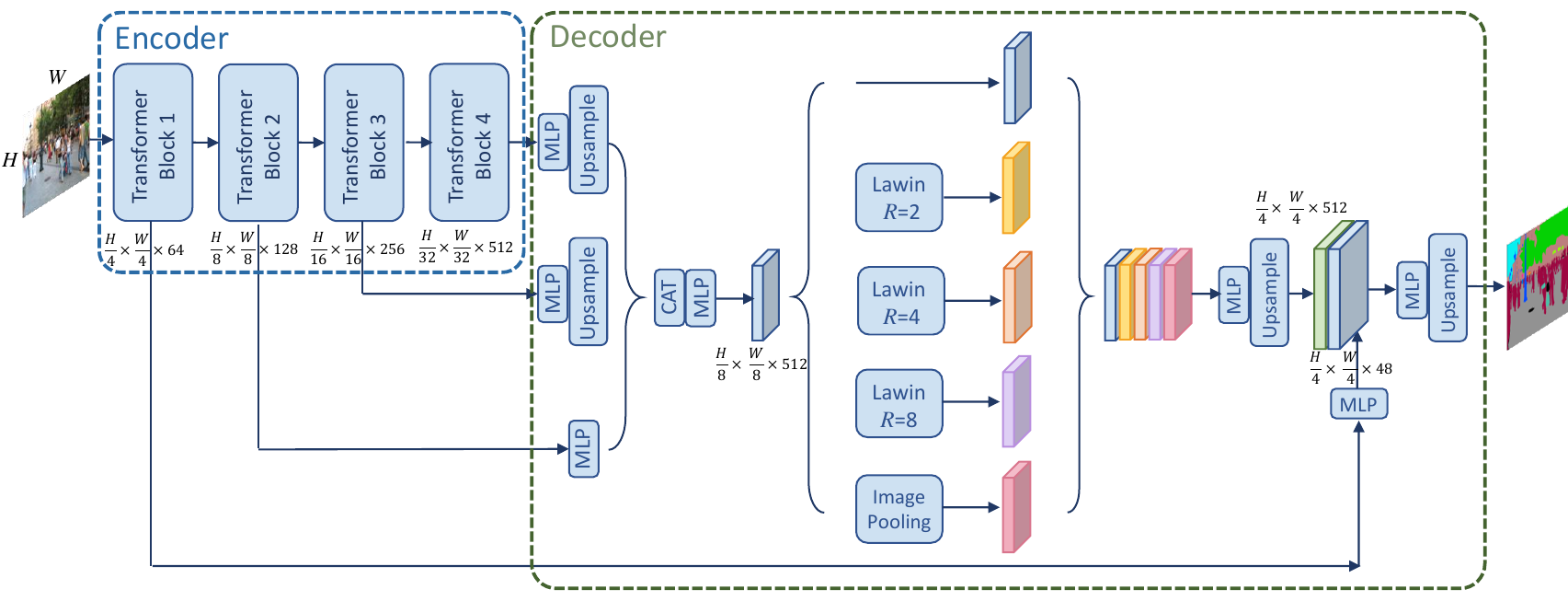} \vspace{-0.8cm}
\end{center}
   \caption{The overall structure of Lawin Transformer. The image is fed into the encoder part,which is a MiT. Then the features from the last three stages are aggregated and fed into the decoder part, which is a LawinASPP. Finally the resulted feature is enhanced with low-level information by the first-stage feature of encoder. "MLP" denotes the multi-layer perceptron. "CAT" denotes concatnating the features. "Lawin" denotes large window attention. "R" denotes the ratio of the size of context patch to query patch.}
\label{fig:lawin transformer}
\end{figure*}
\subsection{Large Window Attention}
Similar with window attention mentioned in section \ref{sec:3.1}, large window attention splits the entire feature map uniformly into several patches. Conversely, when large window attention sliding over the image, the current patch is allowed to query a larger area. For simplicity, we denote the query patch as $\mathbf{Q} \in \mathbb{R} ^ \mathit{P^2 \times C}$ and the queried large context patch as $ \mathbf{C} \in \mathbb{R} ^ \mathit{R^2\times P^2 \times C}$, where $R$ is the ratio of the context patch size to the query patch size, and $P^2$ is the area of patch. Because the computational complexity of attention is $O\left( P^2\right)$, when the spatial size of $ \mathbf{C}$ is increased by $R$ times, the computational complexity increases to $O\left( R^2 P^2\right)$. Under this circumstance, the computation of attention is not limited to the $P\times P$ local patch, and even unaffordable if ratio $R$ or input resolution is very large. To preserve the original computational complexity, we pool $\mathbf{C}$ to an abstract tensor with a downsampling ratio of $R$, reducing the spatial size of context patch back to $\left( P, P\right)$. However, there are certain drawbacks associated with such an easy process. The downsampling of context patch inevitably discards the abundant dependencies between $\mathbf{Q}$ and $\mathbf{C}$ especially as $R$ is large. To mitigate the inattention, we naturally adopt the multi-head mechanism and let the number of head strictly equal to $R^2$, thereby formulating the attention matrix from $\left(P^2, P^2\right)$ to $\left(R^2, P^2, P^2\right)$. It is notable that the number of head has no impact on the computational complexity.

There has been researches revealing that, with certain techniques regularizing the head subspace, multi-head attention can learn desired diverse representations~\cite{child2019generating, cordonnier2019relationship, d2021convit}. Considering that the spatial information becomes abstract after downsampling, we intend to strengthen the spatially representational power of multi-head attention. Motivated by that in MLP-Mixer the token-mixing MLP is complementary to channel-mixing MLP for gathering spatial knowledge, we define a set of head-specific \textit{position-mixing} $\rm{MLP}=\left\{\rm{MLP_{1}}, \rm{MLP_{2}},...,\rm{MLP_{h}} \right\}$. As illustrated in Fig.~\ref{fig:lawin attention}, every head of the pooled context patch is pushed into its corresponding token(position)-mixing MLP, and spatial positions within the same head communicate each other in an identical behavior. We term the resulting context as \textit{position-mixed} context patch and denote it as $\mathbf{C^{\rm{P}}}$, which is calculated by:
\begin{align}
& \mathbf{\hat{C}} = \rm{Reshape}\left(\mathit{h}, \mathit{{C} \slash {h}}, \mathit{P^2} \right)\left( \varphi \left(\mathbf{C} \right) \right),  \label{eq:rel8} \\
& \mathbf{C}_{\rm{h}} = \rm{MLP_{h}} \left(\mathbf{\hat{C}}_{h} \right)+\mathbf{\hat{C}}_{h},  \label{eq:rel9} \\
& \mathbf{C}^{\rm{P}} = \rm{Reshape}\left(\textit{C}, \textit{P}^2 \right)\left( \rm{concat}\left[\mathbf{C}_{1};\mathbf{C}_{2};...;\mathbf{C}_{\rm{h}}\right] \right),\label{eq:rel10}
\end{align}
where $\mathbf{\hat{C}_{\rm{h}}}$ denotes the $h$-th head of $\mathbf{\hat{C}}$ and $\rm{{MLP}_{h}} \in \mathbb{R} ^ {\mathit{P^2 \times P^2 }} $ is the ${h}$-th transformation strengthening the spatial representations for the ${h}$-th head, and $\varphi$ denotes the average pooling operation. With the \textit{position-mixed} context $\mathbf{C}^{\rm{P}}$, we can reformulate the Eq. (\ref{eq:rel3}) and Eq. (\ref{eq:rel4}) as follows:
\begin{align}
& \mathit{A}=\rm{softmax} \left( \frac{ {\left({\mathbf{W_{\rm{q}}}}{\mathbf{Q}_{h}}\right)}{\left({\mathbf{W_{\rm{k}}}}{\mathbf{C}}^{P}_{h}\right)^{\rm{T}}}}{ \sqrt{\mathit{D_{h}}}} \right)\left(\mathbf{W_{\rm{v}}}\mathbf{C}^{P}_{h}\right), \label{eq:rel11} \\
& \rm{MHA}=\rm{concat}\left[\mathit{A_{1}};\mathit{A_{2}};...;\mathit{A_{h}} \right] {\mathbf{W_{\rm{mha}}}}. \label{eq:rel12}
\end{align}

One primary concern is on the overhead of $\rm{MLP}$, so we list the computational complexity of \textit{local window attention} and \textit{large window attention}:
\begin{align}
& \rm{\Omega}\left(Lowin \right)= 4\mathit{\left(HW\right)}\mathit{C^2}+2\mathit{{\left(HW\right)}P^2}\mathit{C}, \label{eq:rel13} \\
& \rm{\Omega}\left(Lawin \right)=4\mathit{\left(HW\right)}\mathit{C^2}+3\mathit{{\left(HW\right)}P^2}\mathit{C},
\label{eq:rel14} 
\end{align}
where $H$ and $W$ are the height and width of entire image respectively, and $P$ is the size of local window. Since $P^2$, usually set to 7 or 8, is much smaller than $C$ in high-level features, the extra expense induced by $\rm{MLP}$ is reasonably neglectable. It is admirable that the computational complexity of large window attention is independent of the ratio $R$.


\subsection{LawinASPP}
To capture multi-scale representations, we adopt the architecture of spatial pyramid pooling (SPP) to collaborate with large window attention and get the novel SPP module called LawinASPP. LawinASPP consists of 5 parallel branches including one shortcut connection, three large window attentions with $R=(2,4,8)$ and an image pooling branch. As shown in Fig.~\ref{fig:lawin transformer}, branches of large window attention provide three hierarchies of receptive fields for the local  window. Following the previous literature on window attention mechanism~\cite{liu2021swin}, we set the patch size of local window to $8$, thus the provided receptive fields are of $\left(16, 32, 64 \right)$. The image pooling branch uses a global pooling layer to obtain the globally contextual information and push it into a linear transformation followed by a bilinearly upsampling opeartion to match the feature dimension. The short path copies the input feature and paste it when all contextual information is output. All resulting features are first concatenated, and a learned linear transformation performs dimensionality reduction for generating the final segmentation map.

\subsection{Lawin Transformer}
Having investigated the advanced HVTs, we select the MiT and Swin-Transformer as the encoder of Lawin Transformer. MiT is designed for serving as encoder of SegFormer~\cite{xie2021segformer} which is a simple, efficient yet powerful semantic segmentation ViT. Swin-Transformer~\cite{liu2021swin} is an extremely successful HVT built upon local window attention. Prior to applying LawinASPP, we concatenate the multi-level features with $output$ $stride=\left(8,16,32 \right)$ by resizing them to the size of feature with $output$ $stride=8$ and use a linear layer to transform the concatenation. The resulting transformed feature with $output$ $stride=8$ is fed into the LawinASPP and then we obtain the feature with multi-scale contextual information. In the state-of-the-art ViT for semantic segmentation, the feature for final prediction of segmentation logits is always derived from 4-level features of encoder. We hence employ the first-level feature with $output$ $stride=4$ to compensate low-level information. The output of LawinASPP is upsampled to the size of a quarter of input image, then fused with the first-level feature by a linear layer. Finally, the segmentation logits are predicted on the low-level-enhanced feature. More details are illustrated in Fig.~\ref{fig:lawin transformer} 

\section{Expriments}
\medskip
\noindent
\textbf{Datasets:}
We conduct experiments on three public datasets including Cityscapes~\cite{cordts2016cityscapes}, ADE20K~\cite{zhou2017scene} and COCO-Stuff~\cite{caesar2018coco}. Cityscapes is an urban scene parsing dataset containing 5,000 fine-annotated images captured from 50 cities with 19 semantic classes. There are 2,975 images divided into training set, 500 images divided into validation set and 1,525 images divided into testing set. ADE20K is one of the most challenging datasets in semantic segmentation. It consists of a training set of 20,210 images with 150 categories, a testing set of 3,352 images and a validation set of 2,000 images. COCO-Stuff is also a very challenging benchmark consists of 164k images with 172 semantic classes. The training set contains 118k images, and the test-dev dataset contains 20k images and the validation set contains of 5k images
\medskip\\
\textbf{Implementation Details:}
Our experiment protocols are exactly the same as those of~\cite{xie2021segformer}. Specially, we use the publicly available ImageNet1K-pretrained MiT~\cite{xie2021segformer} as the encoder of Lawin Transformer. All experiments in this section are implemented based on MMSegmentation~\cite{mmseg2020} codebase on a server with 8 Tesla V100. When doing the ablation study, we choose MiT-B3 as encoder and train all models for 80k iterations. Unless specified, all results are achieved by single-scale inference. Note that all results of other methods are obtained by ours training the official code.
\begin{table*}
  \small
  \centering
  \renewcommand\arraystretch{1.15}
  
  \begin{tabular}{l|c|cc|cc|cc} \hline
  
  \hline

  \small{Dataset} & & \multicolumn{2}{c|}{ADE20K}& \multicolumn{2}{c|}{Cityscapes}& \multicolumn{2}{c}{COCO-Stuff} \\
  \hline
  \small{Method}  & Params(M)$\downarrow$ & FLOPs(G)$\downarrow$ & mIoU(SS/MS)$\uparrow$ & FLOPs(G)$\downarrow$ & mIoU(SS/MS)$\uparrow$ & FLOPs(G)$\downarrow$ & mIoU(SS)$\uparrow$ \\
  \hline
  SegFormer-B0  & 3.8 & 8.4 & 38.1 / 38.6 & 125.5 & 76.5 / 78.2 &  8.4 & 35.7  \\
  Lawin-B0  & 4.1 & 5.3 & \textbf{38.9} / \textbf{39.6}\textbf{} & 99.3 & \textbf{77.2} / \textbf{78.7} & 5.3 &  \textbf{36.2} \\
  \hline
  SegFormer-B1  & 13.7 & 15.9  & 41.7 / 42.8 &  243.7 & 78.5 / 80.0  & 15.9 &  40.2 \\
  Lawin-B1  & 14.1 & 12.7 & \textbf{42.1} / \textbf{43.1}  & 217.5 & \textbf{79.0} / \textbf{80.4} & 12.7  &  \textbf{40.5}  \\
  \hline
  SegFormer-B2 & 27.5 & 62.4 & 46.5 / 47.5 & 717.1 & 81.0 / 82.2 & 62.4 & 44.5\\
  Lawin-B2  & 29.7 & 45.0&  \textbf{47.8} / \textbf{48.8} & 562.8 &  \textbf{81.7} / \textbf{82.7} & 45.0 & \textbf{45.2}\\
  \hline
  SegFormer-B3  & 47.3 & 79.0 & 48.7 / 49.2 & 962.9 &81.7 / 83.3  & 79.0 & 45.4\\
  Lawin-B3  & 49.5 & 61.7 & \textbf{50.3} / \textbf{51.1}& 808.6 &\textbf{82.5} / \textbf{83.7} & 61.7 & \textbf{46.6} \\
  \hline
   SegFormer-B4  & 64.1 & 95.7 & 49.6 / 50.4 & 1240.6 &82.2 / 83.6 & 95.7 & 46.4\\
  Lawin-B4  & 66.3 & 78.2 & \textbf{50.7} / \textbf{51.4} &1086.2 &\textbf{82.7} / \textbf{83.8}&78.2 & \textbf{47.3} \\
  \hline
   SegFormer-B5  & 84.7 & 183.3 & 50.7 / 51.2 & 1460.4 &82.3 / 83.7 & 111.6 & 46.7 \\
  Lawin-B5  & 86.9 & 159.1 & \textbf{52.3} / \textbf{53.0} & 1306.4 & \textbf{82.8} / \textbf{83.9} & 94.2& \textbf{47.5} \\
  \hline

  \hline
  
  \end{tabular}
  \caption{Comparison of SegFormer with Lawin Transformer.}
  \label{tab:seg}
\end{table*}

\begin{table}
  \small
  \centering
  \renewcommand\arraystretch{1.1}
  
  \begin{tabular}{lcccc} \hline
  
  \hline
  Method & FLOPs(G)$\downarrow$ & Params(M)$\downarrow$ & mIoU(SS/MS)$\uparrow$\\
  \hline

   Uper-T & 236.1 &  59.9 & 44.5 / 45.8 \\
   Mask-T & 59.0 & 41.8 & \textbf{46.7 / 48.8} \\ 
   Lawin-T &  \textbf{48.9} &  \textbf{34.5} & 45.3 / 46.9 \\
   \hline
   Uper-S & 259.3 &  81.3  & 47.7 / 49.6 \\
   Mask-S & 79.0 &  63.1  & \textbf{49.8} / \textbf{51.0} \\
   Lawin-S &  \textbf{72.0} &  \textbf{55.9}  & 48.7 / 50.4 \\
   \hline
   $\text{Uper}^{\dagger}\text{-}\text{B}^\star$ &  470.4 & 121.4  & 51.6 / 53.0 \\
   $\text{Mask}^{\dagger}\text{-}\text{B}^\star$ &  195.0 & 101.9  & 52.7 / 53.9 \\
   $\text{Lawin}^{\dagger}\text{-}\text{B}^\star$ &  \textbf{172.9} & \textbf{94.5}  & \textbf{53.0 / 54.3} \\
   \hline
   $\text{Uper}^{\dagger}\text{-}\text{L}^\star$ &  646.4 & 233.4  & 52.7 / 54.1\\
   $\text{Mask}^{\dagger}\text{-}\text{L}^\star$ &  375.0 & 212.0  & 54.1 / 55.6 \\
   $\text{Lawin}^{\dagger}\text{-}\text{L}^\star$ &  \textbf{350.6} & \textbf{201.2}  & \textbf{54.7 / 56.2} \\
  \hline

  \hline
  
  \end{tabular}\vspace{-0.1cm}
  \caption{Comparison of Swin-Lawin Transformer with MaskFormer and Swin-UperNet on ADE20K. The method marked with $\dagger$ takes cropped input of $640\times640$. The method marked with $\star$ indicates that its encoder is pretrained on ImageNet22k.}
  \label{tab:swin}
\end{table}

\begin{table}
  \small
  \centering
  \renewcommand\arraystretch{1.15}
  
  \begin{tabular}{lcc|ccc} \hline
  
  \hline
  \small{Method}  & FLOPs(G) & Params(M) & mIoU \\
  \hline
  PSP~\cite{zhao2017pyramid}  & 48.2 & 49.8 & 47.8 \\
  ASPP~\cite{chen2017rethinking} & 82.9 & 57.0 & 48.5 \\
  SEP-ASPP~\cite{chen2018encoder} & 57.2 & 50.7 & 48.2 \\
  LawinASPP & 61.7 & 49.5 & \textbf{49.9} \\
  \hline

  \hline
  
  \end{tabular}
  \caption{Results of different SPP modules when coupled with MiT-B3 on ADE20K.}
  \label{tb:spp}
\end{table}

\subsection{Comparison with SegFormer}
To demonstrate the improved efficiency of Lawin Transformer, we compare it with SegFormer~\cite{xie2021segformer}. Both of them is built upon window attention and takes MiT as encoder. To enable fairness, we reimplement SegFormer in our environment. Table~\ref{tab:seg} shows the comparison on parameters, FLOPs and mIoU. Apparently, across all variants of MiT (B0$\to$B5), Lawin Transformer trumps SegFormer on mIoU and FLOPs at a little extra parameters. When the light-weight MiT-B0 and MiT-B1 serve as encoder, Lawin Transformer can improve the performance with an adorable saving of computation cost. For example, Lawin-B0 uses much less FLOPs (by $3.1$) to obtain a gain of $0.5\%$ mIoU on COCO-Stuff dataset, and a gain of $0.8\%$ on ADE20K dataset. Moreover, we observe that in some cases, Lawin Transformer can bridge the performance gap caused by the model capacity of encoder. For instance, SegFormer-B3 performs worse than SegFormer-B4 on all three datasets. But if replacing the original decoder with LawinASPP, the resulted Lawin-B3 outperforms SegFormer-B4 by $0.6\%$ mIoU and yields a computation saving of $34$G FLOPs on ADE20K, even using much less parameters. Also, on Cityscapes, Lawin-B4 improves over SegFormer-B5 by $0.4\%$ with nearly a third less computation cost, Lawin-B3 improves over SegFormer-B5 by $0.2\%$ with nearly a half less computation cost and parameters. These empirical results suggest that semantic segmentation ViT could encounter a performance bottleneck as the capacity of encoder continues to increase. In contrast with simply enlarging the encoder, LawinASPP presents a promising and efficient way to overcome the bottleneck by capturing the rich contextual information.

\subsection{Comparison with UperNet and MaskFormer}
To further show the efficiency, we replace MiT with Swin-Transformer~\cite{liu2021swin} and compare the Swin-Lawin Transformer with Swin-UperNet and MaskFormer on ADE20K as shown in table~\ref{tab:swin}. From table~\ref{tab:swin}, we have following observations. Firstly, compared with Swin-UperNet, Swin-Lawin improves the performance largely and saves a great deal of computation cost. In particular, Lawin Transformer with Swin-B can outperform UperNet with Swin-L at nearly a quarter of its computation cost. Secondly, compared with Swin-MaskFormer, Swin-Lawin consistently uses less FLOPs and parameters across all variants of Swin-Transformer. Finally, through a closer look at performance, we find that Swin-Lawin performs worse than MaskFormer when the capacity of encoder is small (Swin-T$\to$S). However, as the capacity of encoder increases (Swin-B$\to$L), Swin-Lawin outperforms MaskFormer. It can be seen that with the increase of capacity, the performance gain created by Swin-Lawin compared with Swin-Uper also become larger. We infer that the short path branch and the low-level information in Lawin Transformer have very important roles in the final prediction (Sec~\ref{sec:4.3.3} discusses the contribution of different hierarchies in Lawin Transformer), which both arise from the multi-level feature of backbone directly. So the more powerful the encoder part, the greater the performance gain from Lawin Transformer.

\subsection{Ablation Study}
\subsubsection{Spatial Pyramid Pooling}
Thanks to the spatial pyramid pooling (SPP)~\cite{lazebnik2006beyond, grauman2005pyramid} architecture in LawinASPP, Lawin Transformer captures multi-scale representations with large window attention in an efficient manner. To study the impact of large window attention and SPP architecture on performance, we choose some representative methods relying on SPP including PPM (Pyramid Pooling Module)~\cite{zhao2017pyramid}, ASPP (Atrous Spatial Pyramid Pooling)~\cite{chen2017rethinking} and SEP-ASPP (Depthwise Separable Atrous Spatial Pyramid Pooling)~\cite{chen2018encoder}. The sharp distinction between LawinASPP with these alternatives is the basic pooling operator. PPM uses pyramid adaptive pooling to capture contextual information at different scales. ASPP uses the atrous convolution to extract multi-scale features. SEP-ASPP uses the depthwise separable atrous convolution~\cite{chollet2017xception} in the place of atrous convolution for the sake of efficiency. Table~\ref{tb:spp} shows parameters, FLOPs and mIoU when MiT-B3 combined with different SPP-based module, which are tested on ADE20K. PPM and SEP-ASPP 
are impressively computational economical, even using less FLOPs than LawinASPP. However there is a considerable performance gap between them and LawinASPP ($2.1\%$ for PPM, $1.7\%$ for SEP-ASPP). ASPP achieves a slightly higher performance than SEP-ASPP, but spends the most computational resources. Through these competitions, LawinASPP proves to be the preferred module to introduce mutli-scale representations into the semantic segmentation ViT, which is mainly attributed to \textit{large window attention}.

\begin{table}
  \small
  \centering
  \renewcommand\arraystretch{1.25}
  
  \begin{tabular}{lccccc} \hline
  
  \hline
 Ratio & Head & C-Mixing & P-Mixing & mIoU  \\
  \hline
  (1,1,1) & (1,1,1) & {\color{red}\XSolidBrush} & {\color{red}\XSolidBrush} & 48.6 \\
   (2,4,8) & (1,1,1) & {\color{red}\XSolidBrush} & {\color{red}\XSolidBrush} & 47.3\\
   (2,4,8) & (4,16,64) & {\color{red}\XSolidBrush} & {\color{red}\XSolidBrush} & 47.9\\
    \hline
 (2,4,8) & (4,16,64) & \textcolor[rgb]{0,0.8,0.3}{\Checkmark} & {\color{red}\XSolidBrush} & 49.1\\
   (2,4,8) & (4,16,64) & {\color{red}\XSolidBrush}&  \textcolor[rgb]{0,0.8,0.3}{\Checkmark}  & \textbf{49.9} \\
    (2,4,8) & (4,16,64) & \textcolor[rgb]{0,0.8,0.3}{\Checkmark}&  \textcolor[rgb]{0,0.8,0.3}{\Checkmark}  & 49.4 \\
  \hline

  \hline
  
  \end{tabular}
  \caption{Results of Lawin-B3 with a variety settings of the ratio, head and MLP type on ADE20K.}
  \label{tab:key}
\end{table}
\begin{figure}[t]
\begin{center}
   \includegraphics[width=0.4\textwidth]{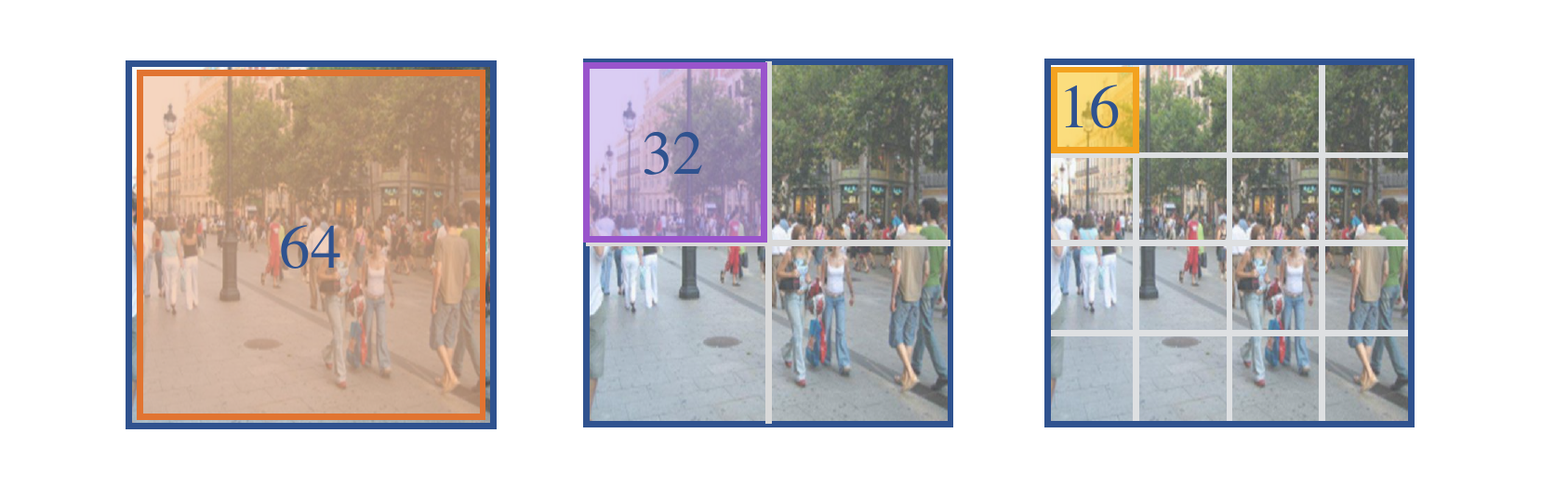} \vspace{-0.7cm}
\end{center}
   \caption{A simple implementation of LawinASPP. The area of both query patch and context patch is set to (64,32,16) }
\label{fig:tweak}
\end{figure}
\begin{table}
  \small
  \centering
  \renewcommand\arraystretch{1.1}
  
  \begin{tabular}{cccc|c} \hline
  
  \hline
 Pooling Ratio & Head & Size & FLOPs(G) & mIoU\\
  \hline

 \hline
   (1,2,4) & (1,4,16) &  16 &  74.6 & \textbf{49.9} \\
   (2,4,8) & (4,16,64) &  8 &  61.7 & \textbf{49.9} \\
  (4,8,16) & (16,64,256) &  4 & \textbf{58.0}  & 49.1 \\
  \hline

  \hline
  
  \end{tabular}\vspace{-0.2cm}
  \caption{Results of Lawin-B3 on ADE20k with context patch of different spatial sizes. "Size" means the spatial size of pooled context.}
  \label{tab:context}
\end{table}

\begin{table}
  \small
  \centering
  \renewcommand\arraystretch{1.25}
  
  \begin{tabular}{cccccc|c} \hline
  
  \hline
$OS$=4 & Short Path & $R$=2 & $R$=4 & $R$=8 & GAP & mIoU\\
  \hline
   \textcolor[rgb]{0,0.8,0.3}{\Checkmark} &\textcolor[rgb]{0,0.8,0.3}{\Checkmark} &   \textcolor[rgb]{0,0.8,0.3}{\Checkmark} & \textcolor[rgb]{0,0.8,0.3}{\Checkmark} & \textcolor[rgb]{0,0.8,0.3}{\Checkmark} & \textcolor[rgb]{0,0.8,0.3}{\Checkmark}& \textbf{49.9} \\
  \textcolor[rgb]{0,0.8,0.3}{\Checkmark}  & \textcolor[rgb]{0,0.8,0.3}{\Checkmark} & {\color{red}\XSolidBrush} & \textcolor[rgb]{0,0.8,0.3}{\Checkmark} & \textcolor[rgb]{0,0.8,0.3}{\Checkmark} &\textcolor[rgb]{0,0.8,0.3}{\Checkmark} & 49.5\\
  \textcolor[rgb]{0,0.8,0.3}{\Checkmark}  & \textcolor[rgb]{0,0.8,0.3}{\Checkmark} & \textcolor[rgb]{0,0.8,0.3}{\Checkmark} & {\color{red}\XSolidBrush} & \textcolor[rgb]{0,0.8,0.3}{\Checkmark} &\textcolor[rgb]{0,0.8,0.3}{\Checkmark} & 49.4\\
  \textcolor[rgb]{0,0.8,0.3}{\Checkmark}  & \textcolor[rgb]{0,0.8,0.3}{\Checkmark} & \textcolor[rgb]{0,0.8,0.3}{\Checkmark} & \textcolor[rgb]{0,0.8,0.3}{\Checkmark} & {\color{red}\XSolidBrush} &\textcolor[rgb]{0,0.8,0.3}{\Checkmark} & 49.4\\
   \textcolor[rgb]{0,0.8,0.3}{\Checkmark} & {\color{red}\XSolidBrush} & \textcolor[rgb]{0,0.8,0.3}{\Checkmark} & \textcolor[rgb]{0,0.8,0.3}{\Checkmark} & \textcolor[rgb]{0,0.8,0.3}{\Checkmark} &\textcolor[rgb]{0,0.8,0.3}{\Checkmark} & 48.9\\
  \textcolor[rgb]{0,0.8,0.3}{\Checkmark}  &  \textcolor[rgb]{0,0.8,0.3}{\Checkmark} & \textcolor[rgb]{0,0.8,0.3}{\Checkmark} & \textcolor[rgb]{0,0.8,0.3}{\Checkmark} & \textcolor[rgb]{0,0.8,0.3}{\Checkmark} &{\color{red}\XSolidBrush} & 49.3\\
 {\color{red}\XSolidBrush}  &  \textcolor[rgb]{0,0.8,0.3}{\Checkmark} & \textcolor[rgb]{0,0.8,0.3}{\Checkmark} & \textcolor[rgb]{0,0.8,0.3}{\Checkmark} & \textcolor[rgb]{0,0.8,0.3}{\Checkmark} &\textcolor[rgb]{0,0.8,0.3}{\Checkmark} & 49.1 \\
  \hline

  \hline
  
  \end{tabular} 
  \caption{Results of Lawin-B3 on ADE20k when different branches are absent.}
  \label{tab:branch} 
\end{table}
\subsubsection{Key Component in Large Window Attention}
\textbf{Pooling and Multi-Head:} Pooling the large context patch with a downsampling ratio $R$ and increasing the head number of MHA to $R^2$ purpose to reducing the computational complexity and recovering the discarded dependencies respectively. To verify the strategy, we implement the first group experiments shown in table~\ref{tab:key}. We first test the performance when the large context patch keeps the spatial size without any downsampling. However, the required memory of this setting is unaffordable so we do a little tweak as illustrated in Fig.~\ref{fig:tweak}, setting the size of query patch equal to context patch. The performance of this simple implementation is lower than the standard implementation by $1.3\%$. If the context patch is pooled to the same size of query patch, the performance degrades severely, only achieving a $47.3\%$ mIoU, owing to the sparsity of attention. Enabling multi-head mechanism can bring an improvement of $0.6\%$, but falling behind the standard and even the simple implementation. This group comparison shows that large window attention with the pooled context patch actually suffers inadequate dependencies, and the multi-head mechanism can alleviate it marginally.
\medskip\\
\noindent
\textbf{Position-Mixing and Channel-Mixing:} In large window attention, we innovatively employ the \textit{position-mixing} operation to strengthen the spatially representational power of multi-head attention. In MLP-Mixer~\cite{tolstikhin2021mlp}, the channel-mixing MLP is applied to learn knowledge of feature channels. That MLP-Mixer uses both kinds of MLPs prompts our investigation of \textit{channel-mixing}. We enhance the communication along feature channels within each head by replacing the token-mixing MLP with channel-mixing MLP. The context patch is downsampled and the multi-head mechansim is enabled. The second group of results listed in table~\ref{tab:key} shows that the channel-mixing MLP boosts the representation of multi-head attention and provides an appreciable performance improvement of $1.2\%$, but not powerful as token-mixing MLP ($2\%$). Furthermore, we make a combination of token-mixing MLP with channel-mixing MLP, like a block in MLP-Mixer, to transform each head subspace along two dimensions, which attains a competitive result of $49.4\%$ mIoU but worse than isolated \textit{position-mixing} ($49.9\%$). With these observations, we argue that the \textit{position-mixing} operation is more useful than \textit{channel-mixing} for recovering the dependencies of spatial downsampling operation. 
\medskip\\
\noindent
\textbf{Spatial Size of Context:} That large window attention pools the context patch to the same spatial size of query patch keeps the balance between efficiency and performance. We are interested in the consequence of disturbing the balance. To be specific, we evaluate the performance in following situations that the context patch is pooled to the spatial size of two times query patch and the context patch is pooled to the half size of query patch. The former sacrifices the computational economy and might be advantageous to performance, and the latter saves more computation cost and might be harmful to performance. It can be found in table~\ref{tab:context} that no apparent performance is obtained in the former case. When the context patch is pooled to a smaller size, the mIoU drops $0.8\%$ and only saves a little computation cost of $3.7$G. Pooling the context patch to the size of query patch is a sensible choice keeping the balance well.

\begin{table}

  \small
  \centering
  \renewcommand\arraystretch{1.15}
  
  \begin{tabular}{lccccc} \hline 
  
  \hline
  \small{Method} & Backbone & FLOPs(G)$\downarrow$ & MS(\%)$\uparrow$ \\
  \hline
  PSPNet~\cite{zhao2017pyramid}  & ResNet101 & 2049 & 80.0 \\
   GCNet~\cite{cao2019gcnet}  & ResNet101 & 2203 & 80.7 \\
   PSANet~\cite{zhao2018psanet}  & ResNet101 & 2178 & 80.9 \\
   NonLocal~\cite{wang2018non}  & ResNet101 & 2224 & 80.9 \\
   DeeplabV3~\cite{chen2017rethinking} & ResNet101 & 2781 & 80.8 \\
   CCNet~\cite{huang2019ccnet}  & ResNet101 & 2225  & 80.7 \\
   DANet~\cite{fu2019dual}  & ResNet101 & 2221 & 82.0 \\
   DNL~\cite{yin2020disentangled}  & ResNet101 & 2224 & 80.7 \\
   OCNet~\cite{yuan2018ocnet} & ResNet101 & 1820 & 81.6 \\
   DeeplabV3+~\cite{chen2018encoder}  & ResNet101 & 2032 & 82.2 \\
   \hline
   SETR-PUP~\cite{zheng2021rethinking}  & $\text{ViT-L}^{\star}$ & --- & 82.2 \\
   Segmenter~\cite{strudel2021segmenter} & $\text{VIT-L}^{\star}$ & --- & 81.3 \\
   SegFormer~\cite{xie2021segformer}  & MiT-B5 & 1460 & 83.5 \\
   $\text{SegFormer}^{\dagger}$ & MiT-B5 & 1460 & 83.7 \\
   Lawin & MiT-B5 & 1306 & 83.7 \\
   $\text{Lawin}^{\dagger}$ & MiT-B5 & 1306 & 83.9 \\
   Lawin  & $\text{Swin-L}^{\star}$ & 1797 & 84.2\\
   $\text{Lawin}^{\dagger}$  & $\text{Swin-L}^{\star}$ & 1797 & \textbf{84.4}\\
   
  \hline

  \hline
  
  \end{tabular} 
  \caption{Performance Comparison on Cityscapes. The backbone marked with $\star$ indicates that it is pretrained on ImageNet22K. The method marked with $\dagger$ takes cropped input of $1024\times1024$.   }
  \label{tb:city}
\end{table}

\begin{table}
  \small
  \centering
  \renewcommand\arraystretch{1.15}
  
  \begin{tabular}{lccccc} \hline 
  
  \hline
  \small{Method} & Backbone & FLOPs(G)$\downarrow$ & MS(\%)$\uparrow$ \\
  \hline
  PSPNet~\cite{zhao2017pyramid}  & ResNet101 & 256 & 45.4\\
   GCNet~\cite{cao2019gcnet}  & ResNet101 & 276  & 45.2 \\
   PSANet~\cite{zhao2018psanet}  & ResNet101 & 272 & 45.4 \\
   NonLocal~\cite{wang2018non}  & ResNet101 & 278 & 45.8 \\
   DeeplabV3~\cite{chen2017rethinking} &ResNet101 & 348 & 46.7 \\
   CCNet~\cite{huang2019ccnet}  & ResNet101 & 278 & 45.0 \\
   DANet~\cite{fu2019dual}  & ResNet101 & 278 & 45.0 \\
   DNL~\cite{yin2020disentangled}  & ResNet101 & 278 & 45.8 \\
   OCNet~\cite{yuan2018ocnet}  & ResNet101 & 227 & 45.4 \\
   DeeplabV3+~\cite{chen2018encoder}  & ResNet101 & 255 &46.4 \\
   OCRNet~\cite{yuan2020object} & HRNetW48 & 165& 44.9\\
   \hline
   $\text{SETR-MLA}^{\dagger}$\text{~\cite{zheng2021rethinking}}  &$\text{ViT-L}^\star$ & --- & 50.3 \\
   $\text{Segmenter}^{\dagger}$\text{~\cite{strudel2021segmenter}}  & $\text{ViT-L}^\star$ & --- & 53.6\\
   $\text{SegFormer}^{\dagger}$\text{~\cite{xie2021segformer}}   & MiT-B5 & 183 & 51.2\\
   $\text{Lawin}^{\dagger}$   & MiT-B5 & \textbf{159} & 53.0 \\
   $\text{Swin-Uper}^{\dagger}$\text{~\cite{liu2021swin}} & $\text{Swin-L}^{\star}$ & 646 & 54.1 \\ 
   $\text{MaskFormer}^{\dagger}$\text{~\cite{cheng2021maskformer}}  & $\text{Swin-L}^{\star}$ & 375 & 55.6 \\
   $\text{Lawin}^{\dagger}$   & $\text{Swin-L}^{\star}$ & 351 & \textbf{56.2}\\
  \hline

  \hline
  
  \end{tabular}
  \caption{Performance Comparison on ADE20K. The backbone marked with $\star$ indicates that it is pretrained on ImageNet22K. The method marked with $\dagger$ takes cropped input of $640\times640$.}
  \label{tab:ade20k}
\end{table}

\subsubsection{Branch in LawinASPP}
\label{sec:4.3.3}
As illustrated in Fig.~\ref{fig:lawin transformer}, LawinASPP aggregates features derived from five branches to gather rich contextual information at multiple scales. Following the aggregation, the first-level feature comes to enhance it with low-level information via an auxiliary branch. We here study the efficacy of the six branches in LawinASPP. In table~\ref{tab:branch}, we report the results when different branches are absent. For the branches of large window attention, the performance drops $0.4\%$, $0.5\%$ and $0.5\%$ for removing the branch with $R=2$, $R=4$ and $R=8$ respectively. The image pooling branch yields an improvement of $0.6\%$ so the global contextual information is an essential hierarchy of LawinASPP. The short path is also indispensable to LawinASPP in that the biggest performance gain of ($1.0\%$) is from this branch. We surprisingly observe that adding the auxiliary branch leads to an improvement of $0.8\%$, which manifests the importance of low-level information.

\begin{table}
  \small
  \centering
  \renewcommand\arraystretch{1.15}
  
  \begin{tabular}{lccccc} \hline 
  
  \hline
  \small{Method} & Backbone & FLOPs(G)$\downarrow$ & mIoU($\%$)$\uparrow$ \\
  \hline
   NonLocal~\cite{wang2018non}  & ResNet101 & 278 & 37.9\\
   DeeplabV3+~\cite{chen2018encoder}  & ResNet101 & 255  & 38.4 \\
   OCRNet~\cite{yuan2020object}  & HRNetW48 & 165 & 42.3 \\
   \hline
   SETR-MLA~\cite{zheng2021rethinking}  & $\text{ViT-L}^\star$ & --- & 45.8 \\
   SegFormer~\cite{xie2021segformer}  & MiT-B5 & 112 & 46.7\\
   Lawin  & MiT-B5 & \textbf{94} & \textbf{47.5}\\
  \hline

  \hline
  
  \end{tabular}
  \caption{Performance Comparion on COCO-Stuff. The backbone with superscript $\star$ indicates that it is pretrained on ImageNet22K.}
  \label{tab:coco}
\end{table}

\subsection{Comprasion with State-of-the-Art}
Finally, we compare our results with existing approaches on the ADE20K, Cityscapes and COCO-Stuff datasets. 

Table \ref{tb:city} shows the results of state-of-the-art methods on Cityscapes dataset. The first group contains the CNN-based semantic segmentation method and the second group contains the ViT-based semantic segmentation method. If not specified, the crop size of input image is $768/769\times768/769$. To boost the performance of Lawin Transformer, we use MiT-B5 and Swin-L as the encoder. Lawin Transformer with Swin-L achieves the best performance on Cityscapes. 

Table \ref{tab:ade20k} reports the performance of state-of-the-art methods on ADE20K dataset. The results are still grouped into two parts consists of CNN-based methods and ViT-based methods. If not specified, the crop size of input image is $512\times512$. Lawin Transformer with Swin-L outperforms all other methods. Lawin Transformer with MiT-B5 uses the least FLOPs ($159$ GFLOPs) and achieves an excellent performance ($53.0\%$ mIoU). 

Table \ref{tab:coco} lists some results of state-of-the-art methods on COCO-stuff. Since there are few paper reporting the performance on COCO-Stuff, we just list the result of representative CNN-based methods. Lawin-B5 obtains the best mIoU of $47.5\%$ and also uses the least FLOPs of $94$G.

\section{Conclusion}
In this work, we develop an efficient semantic segmentation transformer called Lawin Transformer. The decoder part of Lawin Transformer is capable of capturing rich contextual information at multiple scales, which is established on our proposed \textit{large window attention}. Compared to the existing efficient semantic segmentation Transformer, Lawin Transformer can achieve higher performance with less computational expense. Finally, we conduct experiments on Cityscapes, ADE20K and COCO-Stuff dataset, yielding state-of-the-art results on these benchmarks. We hope Lawin Transformer will inspire the creativity of semantic segmentation ViT in the future.
{\small
\bibliographystyle{ieee_fullname}
\bibliography{egbib}
}

\end{document}